%% file: neurips_2025.tex
\documentclass{article}

\usepackage[preprint]{neurips_2025}

\usepackage[utf8]{inputenc} 
\usepackage[T1]{fontenc}    
\usepackage{hyperref}       
\usepackage{url}            
\usepackage{booktabs}       
\usepackage{amsfonts}       
\usepackage{nicefrac}       
\usepackage{microtype}      
\usepackage{xcolor}         

\usepackage{graphicx}
\usepackage{amsmath}
\usepackage{amssymb}
\usepackage{booktabs}
\usepackage{bbm}
\usepackage{subfigure}
\usepackage{multirow}
\usepackage{makecell}

\usepackage{wrapfig}    
\usepackage{caption}    
\usepackage{lipsum}     

\usepackage{bbding}
\usepackage{pifont}
\usepackage{wasysym}
\usepackage{amssymb}
\usepackage[flushleft]{threeparttable}

\newcommand{\cmark}{\ding{51}}%
\newcommand{\xmark}{\ding{55}}%

\usepackage{subcaption} 
\usepackage{booktabs} 

\usepackage{enumitem}

\title{Exploiting Unlabeled Structures through Task Consistency Training for Versatile Medical Image Segmentation}

\author{%
  Shengqian Zhu, Jiafei Wu, Xiaogang Xu, Chengrong Yu, \AND Ying Song, Zhang Yi, Guangjun Li, Junjie Hu \\
  Sichuan University \\
  \texttt{2023323040021@stu.scu.edu.cn} \\
}

\author{%
  Shengqian Zhu\textsuperscript{\rm 1}, Jiafei Wu\textsuperscript{\rm 2}, Xiaogang Xu\textsuperscript{\rm 3}, Chengrong Yu\textsuperscript{\rm 1}, \AND Ying Song\textsuperscript{\rm 1}, Zhang Yi\textsuperscript{\rm 1}, Guangjun Li\textsuperscript{\rm 1}, Junjie Hu\textsuperscript{\rm 1}  \\
  \textsuperscript{\rm 1} Sichuan University\\
  \textsuperscript{\rm 2} University of Hong Kong,
  \textsuperscript{\rm 3} The Chinese University of Hong Kong \\
  \texttt{2023323040021@stu.scu.edu.cn} \\
}

\begin{document}

\maketitle

\input{sec/0_abstract}

\input{sec/1_intro}

\input{sec/2_related_work}

\input{sec/3_method}

\input{sec/4_experiments}

\input{sec/5_conclusion}

\newpage
{
    \small
    \bibliographystyle{plain}
    \bibliography{main}
}

\newpage
\appendix

\input{sec/X_suppl}

\end{document}

%% file: sec/0_abstract.tex
\begin{abstract}

Versatile medical image segmentation (VMIS) targets the segmentation of multiple classes, while obtaining full annotations for all classes is often impractical due to the time and labor required. Leveraging partially labeled datasets (PLDs) presents a promising alternative; however, current VMIS approaches face significant class imbalance due to the unequal category distribution in PLDs. Existing methods attempt to address this by generating pseudo-full labels. Nevertheless, these typically require additional models and often result in potential performance degradation from label noise. In this work, we introduce a Task Consistency Training (TCT) framework to address class imbalance without requiring extra models. TCT includes a backbone network with a main segmentation head (MSH) for multi-channel predictions and multiple auxiliary task heads (ATHs) for task-specific predictions. By enforcing a consistency constraint between the MSH and ATH predictions, TCT effectively utilizes unlabeled anatomical structures. To avoid error propagation from low-consistency, potentially noisy data, we propose a filtering strategy to exclude such data. Additionally, we introduce a unified auxiliary uncertainty-weighted loss (UAUWL) to mitigate segmentation quality declines caused by the dominance of specific tasks. Extensive experiments on eight abdominal datasets from diverse clinical sites demonstrate our approach's effectiveness.

\end{abstract}

%% file: sec/1_intro.tex
\section{Introduction}

\begin{figure}[t]
  \centering
   \includegraphics[width=\linewidth]{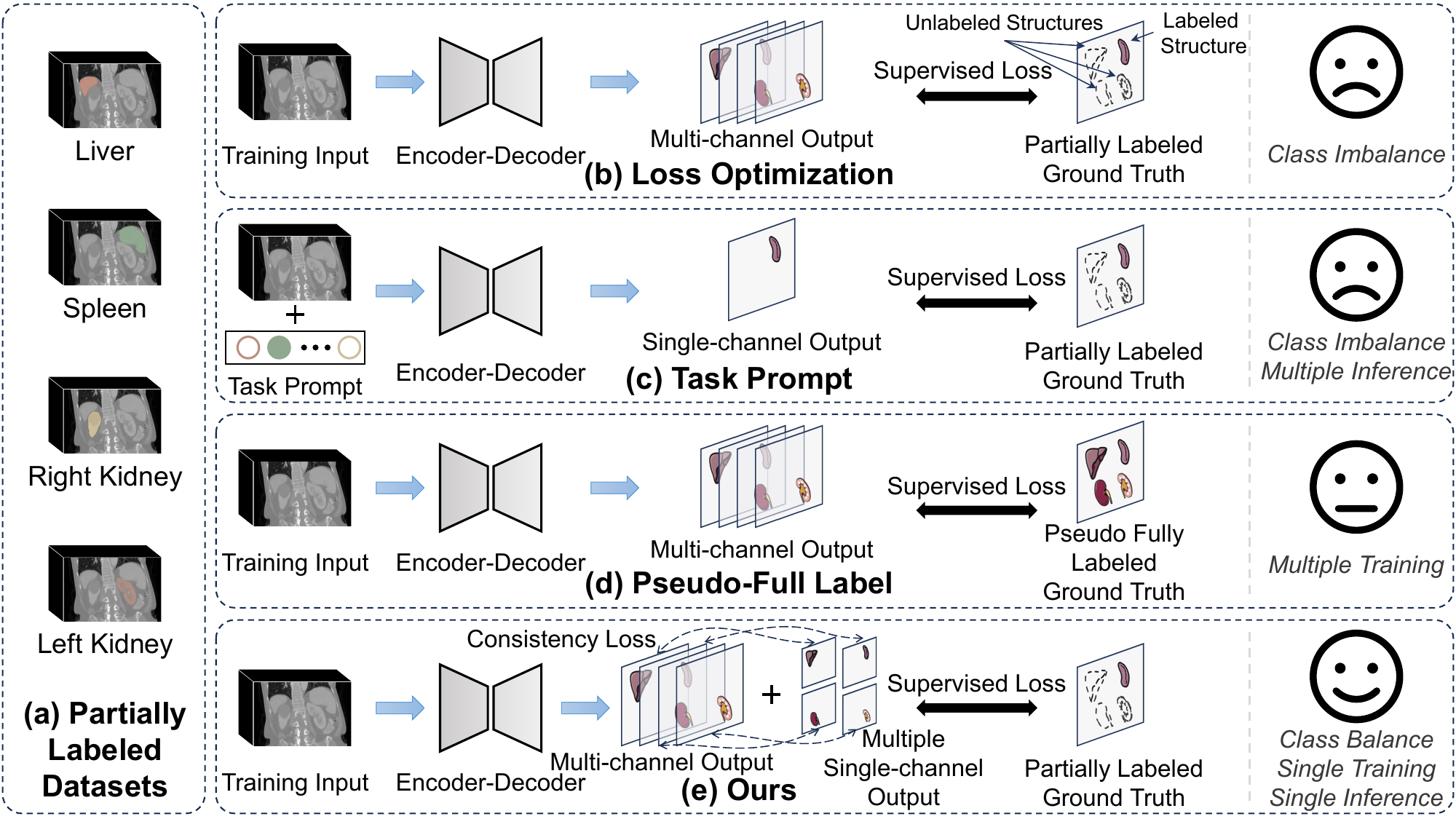}
   \vspace{-0.2in}
   \caption{(a) Illustration of PLDs, including liver, spleen, right kidney, and left kidney. This task aims to train a versatile model on multiple PLDs. The conceptual comparison among (b) loss optimization models, (c) task prompt models, (d) pseudo-full label models, and (e) our approach.}
   \label{fig:first_graph}
   \vspace{-0.2in}
\end{figure}

Accurate and robust segmentation of target organs and tissues is essential for numerous clinical applications, including treatment planning~\cite{bauer2013survey, burgos2017iterative, hu2021incorporating} and prognosis evaluation~\cite{zhang2020clinically}. Versatile medical image segmentation (VMIS) models have garnered significant research interest due to their streamlined workflow, scalability, and efficient resource use. However, developing such models typically requires full labeling of all targets, which is often impractical due to privacy concerns and the high costs associated with pixel-wise annotations~\cite{hesamian2019deep}. To address these challenges, recent studies have focused on leveraging partially labeled datasets (PLDs, see Fig.~\ref{fig:first_graph}a) from diverse clinical sites, using varied scanners and protocols, as an alternative to fully labeled datasets (FLDs).

Partially supervised learning (PSL)~\cite{zhou2019prior} has emerged as a prominent area of research, addressing the challenge where each training example is associated with a set of candidate labels, only some of which are ground truth. Current versatile segmentation methods using PSL generally fall into two categories based on the supervision type: 1) partial label methods and 2) pseudo-full label strategies. Partial label methods use carefully designed loss functions\cite{fang2020multi, shi2021marginal} (Fig.~\ref{fig:first_graph}b) or task-specific priors~\cite{zhang2021dodnet, liu2023clip, ye2023uniseg, liu2023ccq, gao2024training} (Fig.~\ref{fig:first_graph}c) to address conflicts between optimization objectives and partially labeled data. However, class imbalance due to unequal category distributions in PLDs often leads to model performance declines, as models are prone to overfitting (Fig.~\ref{fig:dataset_a} MSD-Spleen). To mitigate this, pseudo-full label approaches~\cite{huang2020multi, feng2021ms, liu2022universal, liu2024cosst} (Fig.~\ref{fig:first_graph}d) generate pseudo labels for unannotated structures in each example, typically using multiple task-specific segmentation models. This approach enables the model to leverage unlabeled anatomical data and supports flexibility in network architecture and loss function selection. However, pseudo-full label methods are hindered by high training costs and often experience performance degradation from noisy labels, especially when training data is limited.

In this paper, we introduce a task consistency training (TCT) framework that mitigates class imbalance without requiring additional training. TCT is structured with a backbone network, a main segmentation head (MSH), and multiple auxiliary task heads (ATHs). The MSH generates multi-channel predictions for all classes, while the ATHs produce task-specific predictions, as shown in Fig.~\ref{fig:main_architecture}. To leverage unlabeled anatomical structures within each partially labeled example, we enforce consistency between the MSH and ATH predictions. This consistency constraint helps sustain segmentation performance for specific categories, even with limited data, by encouraging aligned predictions for the same class across MSH and ATHs. Through this approach, TCT effectively utilizes unannotated structures to reduce class imbalance.

Low consistency between the MSH and ATHs predictions can introduce noise, leading to error accumulation and propagation. To address this, we propose a consistency filtering strategy that retains only high-consistency information. Specifically, we use the Intersection over Union (IoU) metric to assess consistency and exclude data that falls below a defined threshold. Additionally, we introduce a unified auxiliary uncertainty-weighted loss (UAUWL) to balance the main and auxiliary tasks. UAUWL assigns learnable uncertainties to both the MSH and ATHs, dynamically adjusting their loss weights based on these uncertainties. This approach effectively avoids segmentation performance degradation by preventing any single task from excessively dominating others.

The contributions of this work lie in three aspects:

\begin{itemize}
  \item 
  We propose a task consistency training framework for VMIS that mitigates class imbalance without the need for additional training.
  
  \item 
  We propose a consistency filtering strategy to exclude low-consistency data, preventing error accumulation and propagation. Additionally, we introduce a multi-task loss function, UAUWL, to enhance the accuracy of TCT.
    
  \item 
  We conduct rigorous experiments on eight abdominal datasets from diverse clinical sites, totaling 1,133 cases. The results demonstrate that our approach outperforms state-of-the-art (SOTA) methods (\textit{e.g.}, with +0.57\% DSC improvement on the CT PLDs).
\end{itemize}

%% file: sec/2_related_work.tex
\section{Related Work}

\subsection{Partially Labeled Versatile Segmentation}

Versatile image segmentation seeks to segment multiple classes of interest with a unified model. Due to the scarcity of FLDs, the current approaches~\cite{chen2019med3d, zhou2019prior, dmitriev2019learning, ulrich2023multitalent, zhou2023partially, fang2020multi, shi2021marginal, zhang2021dodnet, liu2023clip, ye2023uniseg, liu2023ccq, huang2020multi, feng2021ms, liu2022universal, liu2024cosst, gao2024training, chen2024versatile} typically train the model using PLDs from various clinical sites. These methods can be broadly categorized into two types based on the supervision signals: partial label and pseudo-full label methods.

\paragraph{Partial Label Models.}
Partial label models have primarily addressed the conflict between multiple optimization objectives and partially labeled data through loss optimization~\cite{fang2020multi, shi2021marginal} or task prompts~\cite{zhang2021dodnet, liu2023clip, ye2023uniseg, liu2023ccq, gao2024training}. For instance, Fang~\cite{fang2020multi} introduced a target-adaptive loss (TAL), which treats pixels lacking ground truth as background, ensuring consistency between multi-channel predictions and partially annotated ground truth. Alternatively, some methods~\cite{zhang2021dodnet, ye2023uniseg, liu2023ccq, gao2024training} use task priors to generate single-channel predictions for each task, promoting versatile segmentation and aligning the predictions with partially labeled ground truth. However, task prompt models~\cite{zhang2021dodnet, ye2023uniseg, liu2023ccq, gao2024training} require multiple inferences to capture all targets. Despite these advances, these models often overlook the class imbalance issue inherent in PLDs.

\paragraph{Pseudo-Full Label Models.}
Several efforts~\cite{huang2020multi, feng2021ms, liu2022universal, liu2024cosst} have aimed to address class imbalance by generating pseudo labels for unannotated structures. Huang \textit{et al.}~\cite{huang2020multi} created a fully annotated dataset with pseudo labels using multiple pre-trained single-organ models and proposed a co-training framework based on pseudo-full labels to cross-supervise a versatile model. Similarly, Feng \textit{et al.}~\cite{feng2021ms} employed multiple single-task teacher models for knowledge distillation to supervise a single versatile student model. Liu \textit{et al.}~\cite{liu2024cosst} introduced an iterative self-training strategy to reduce noise in pseudo labels and improve model performance. While pseudo-full labels help alleviate class imbalance, they come with significantly higher training costs and complexity during testing. In contrast, our approach resolves class imbalance without additional training processes and enables obtaining all target predictions in a single forward pass during inference.

\subsection{Consistency in Semi-Supervised Segmentation}
Semi-supervised segmentation methods aim to better utilize unlabeled data, with consistency learning being a widely used strategy. The core idea is to encourage the model to produce consistent outputs for the same unlabeled data under different transformations, thereby improving generalization. For example, Ouali \textit{et al.}~\cite{ouali2020semi} introduced noise into intermediate features and enforced pixel consistency between the main and auxiliary decoders' predictions. In addition to data-level consistency, Luo \textit{et al.}~\cite{luo2021semi} established task-level consistency between a set-level regression task and a pixel-level classification task to leverage unlabeled data. These methods~\cite{zamir2020robust, luo2021semi} focus on building consistency across different tasks in semi-supervised settings. In this work, we explore task consistency as a means to mitigate the class imbalance issue in VMIS.

%% file: sec/3_method.tex
\begin{figure}[t]
  \centering
   \includegraphics[width=1.0\linewidth]{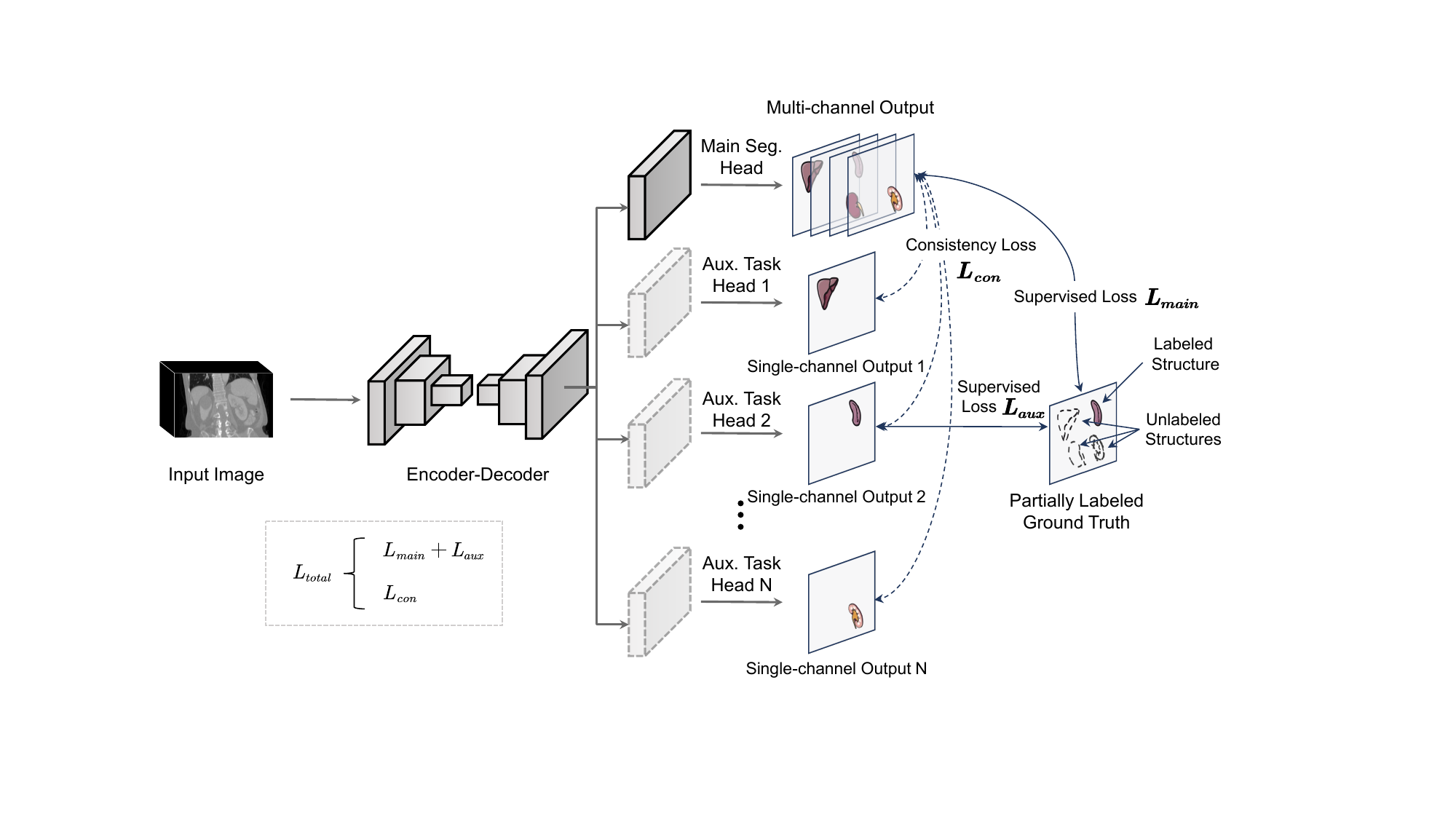}
   \vspace{-0.2in}
   \caption{Overview of our TCT framework. Seg. and Aux. denote segmentation and auxiliary, respectively.
   The overall loss function consists of $\mathcal{L}_{\mathrm{main}}$ (Eq.~\ref{eqn:main_sup}), $\mathcal{L}_{\mathrm{aux}}$ (Eq.~\ref{eqn:aux_sup}), and $\mathcal{L}_{\mathrm{con}}$ (Eq.~\ref{eqn:con}). In the training stage, MSH generates multi-channel output for all $N$ classes. For each of the $N$ classes, the corresponding ATH produces a single-channel prediction. In the inference stage, these $N$ ATHs are not used for the forward pass, reducing the computational load.}
   \label{fig:main_architecture}
   \vspace{-0.2in}
\end{figure}

\section{Method}

\subsection{Problem Definition}

Due to the scarcity of FLDs, the current versatile models are trained on the PLDs. Without loss of generality, let us consider a scenario involving a total of $N$ classes of interest, with their label index set denoted as $\Omega = \{1, 2, \cdots, N \}$, where $ |\Omega| = N$. Given a PLD containing $M$ classes of interest, the label index set is represented by $\Phi$, where $\Phi \subset \Omega$ and $|\Phi| = M$. For each partially labeled sample, there is an associated unannotated label index set $\Phi^{\complement}$, where $\Phi^{\complement} = \Omega \setminus \Phi = \{c\in \Omega \mid c\notin \Phi\}$. The objective is to train a versatile model capable of segmenting $N$ classes.

\subsection{Overview}
As discussed in the introduction, our goal is to leverage unlabeled structures to mitigate the class imbalance issue. The proposed TCT framework (Fig.~\ref{fig:main_architecture}) consists of a main segmentation head (MSH) and a set of $N$ auxiliary task heads (ATHs) atop a 3D-UNet backbone, with one head using a convolutional layer. The MSH generates multi-channel predictions for all classes of interest $\Omega$, while the ATHs produce task-specific predictions for each class in $\Phi$. We enforce consistency between the MSH and ATHs predictions for the same class. To avoid noise from low-consistency data, which can lead to error accumulation and propagation, we introduce a consistency filtering strategy to retain only high-consistency data. By doing so, the unlabeled structures $\Phi^{\complement}$ within partially labeled samples are effectively utilized during training, promoting comprehensive learning of the segmentation network and helping to address the class imbalance issue.

For each partially labeled example, it is crucial to utilize the ground truth supervision from the labeled structures while also exploring the underlying information in the unlabeled structures. The labeled structures are used to train the MSH and $N$ ATHs in a supervised manner, while the unlabeled structures are leveraged via our proposed TCT.

\subsection{Supervised Training from Ground Truth}
Given the presence of only partial labels, applying conventional fully supervised learning losses directly could lead to incompatibility between the model outputs and the supervision signals. To handle this challenge, we adopt the pioneering conceptualization in the literature~\cite{fang2020multi}, which treats unlabeled structures as background and merges their corresponding model output probabilities into the background channel. Let $p_i^n$ and $y_i^m$ denote the predicted probability output by the MSH and the ground truth of the partially labeled samples, respectively. In this context, the subscript $i$ refers to the voxel index, and the superscripts $n$ and $m$ indicate the indices of the channels associated with $i$-th voxel, where $n \in \{0\} \cup \Omega$ and $m \in \{0\} \cup \Phi$. Note that `0' represents the background channel. The MSH is trained using Dice loss~\cite{abdollahi2020vnet}, as
\begin{equation}
\begin{aligned}
    \mathcal{L}_{\mathrm{main}} = 1-\frac{1}{M+1}\sum_{c\in\{0\} \cup \Phi}\frac{2 {\textstyle \sum_{i}^{N_v}} q_{i}^{c} y_{i}^{c}}{ {\textstyle \sum_{i}^{N_v}} q_{i}^{c} +  {\textstyle \sum_{i}^{N_v}} y_{i}^{c}},
    \label{eqn:main_sup}
\end{aligned}
\end{equation} where $N_v$ denotes the total number of voxels in the input. 
The merged prediction probability $q_{i}^{c}$ is calculated as
\begin{equation}
\begin{aligned}
    q_{i}^{c} = \left\{
   \begin{aligned}
    & {\textstyle \sum_{n \in \{0\} \cup \Phi^{\complement}}} p_{i}^{n}   & & \textrm{if } c = 0 \\
    & p_{i}^{c}                          & & \textrm{if } c \in \Phi \\
   \end{aligned}
    \right..
    \label{eqn:main_merge}
\end{aligned}
\end{equation} 
Similarly, an incompatibility arises between the outputs generated by the ATHs and the corresponding supervision signals when the training sample contains more than one labeled anatomical structure. We define $g_{i}^{j}$ ($j \in \Omega$) as the predicted probability of the $i$-th voxel produced by the $j$-th ATH. The ATHs are supervised by the Dice loss as
\begin{equation}
\begin{aligned}
    \mathcal{L}_{\mathrm{aux}} = \sum_{j \in \Omega} (1-\frac{1}{2}\sum_{c\in\{0, j\}}\frac{2 {\textstyle \sum_{i}^{N_v}} g_{i}^{c} z_{i}^{c}}{ {\textstyle \sum_{i}^{N_v}} g_{i}^{c} +  {\textstyle \sum_{i}^{N_v}} z_{i}^{c}}),
    \label{eqn:aux_sup}
\end{aligned}
\end{equation} where $z_{i}^{c}$ represents the supervision signals merged from ground truth $y_{i}^{c}$ as follows
\begin{equation}
\begin{aligned}
    z_{i}^{c} = \left\{
   \begin{aligned}
    & {\textstyle \sum_{m \in \{0\} \cup (\Phi \setminus \{j\} )}} y_{i}^{m}   & & \textrm{if } c = 0 \\
    & y_{i}^{c}                          & & \textrm{if } c \in \Omega \\
   \end{aligned}
    \right..
    \label{eqn:aux_merge}
\end{aligned}
\end{equation}

\subsection{Task Consistency Training}

\paragraph{Objective.}
As discussed earlier, we enforce consistency between the predictions of the MSH and ATHs. Specifically, we minimize the difference between $p_i^c$ and $g_i^c$ for the same anatomical structure in each partially labeled sample. Here, $p_i^c$ and $g_i^c$ denote the predicted probabilities for class $c$ output by the MSH and ATHs, respectively. We use mean squared error to measure the distance between $p_i^c$ and $g_i^c$, aiming to minimize this distance as our training objective.

\paragraph{Consistency Filtering.}
Low-consistency data between the MSH and ATHs may contain noise, leading to error accumulation and propagation. To mitigate this, we propose a consistency filtering strategy to exclude low-consistency data. We use the Intersection over Union (IoU) metric to assess consistency between $p_i^c$ and $g_i^c$, retaining only the loss terms with an IoU greater than or equal to the threshold $\theta$, which are then back-propagated.

\paragraph{Implementation.}
Before defining the final task consistency loss, it is important to note the channel-level incompatibility between $p_i^c$ and $g_i^c$. To address this, we transform $p_i^c$ into $q_i^c$ using the procedure outlined in Eq.~\ref{eqn:main_merge}, as

\begin{equation}
\begin{aligned}
    q_{i}^{c} = \left\{
   \begin{aligned}
    & {\textstyle \sum_{n \in \{0\} \cup \Phi^{\complement}}} p_{i}^{n}   & & \textrm{if } c = 0 \\
    & p_{i}^{c}                          & & \textrm{if } c \in \Omega \\
   \end{aligned}
    \right..
    \label{eqn:con_merge}
\end{aligned}
\end{equation} 
Thereafter, we define the task consistency loss $\mathcal{L}_{\mathrm{con}}$ by using the mean squared error as

\begin{equation}
\begin{aligned}
    \hspace{-2mm}
    \mathcal{L}_{\mathrm{con}}\!=\!\sum_{j \in \Omega} \mathbbm{1}_{[IoU(q^{j}, g^{j}) \ge \theta]} (\frac{1}{2 N_v}\!\sum_{c \in \{0, j \}}\!\sum_{i}^{N_v} (q_{i}^{c} - g_{i}^{c})^2),
    \label{eqn:con}
\end{aligned}
\end{equation} 
where $\mathbbm{1}$ represents the indicator function. The $IoU(q^{j}, g^{j})$ computes the IoU value between the MSH and ATH predictions for class $j$. To mitigate instability from outliers, we set $\theta$ as the median of the IoU values computed across all ATHs. The task consistency loss $\mathcal{L}_{\mathrm{con}}$ leverages ATHs to extract unannotated structure information, which is then used to enhance the learning of the MSH.

\subsection{Overall Loss Function}
\paragraph{Unified Auxiliary Uncertainty-Weighted Loss.}
To balance the main segmentation task with multiple auxiliary tasks, we use the commonly applied multi-task uncertainty-weighted loss (UWL) function~\cite{kendall2018multi, xu2022mtformer}. UWL assigns an uncertainty value $\sigma_{i}$ to each task and adjusts the loss weight dynamically based on these uncertainties. However, directly applying UWL in TCT and assigning individual uncertainties to each ATH can lead to excessive dominance by certain tasks, hindering overall segmentation efficiency. Since segmentation difficulty varies across tasks, ATHs with lower uncertainty values tend to dominate, limiting the performance of others. This over-dominance is particularly evident in cases of class imbalance, where category scales differ significantly. To address this, we introduce a unified auxiliary uncertainty-weighted loss (UAUWL), assigning a single uncertainty value to all auxiliary tasks. UAUWL effectively mitigates the negative impact of over-dominance on segmentation performance.

\paragraph{Overall Loss.}

The overall loss function is defined as

\begin{equation}
\begin{aligned}
    \hspace{-2mm}
    \mathcal{L}_{total}\!=\!\frac{1}{\sigma_{1}^{2}} \mathcal{L}_{\mathrm{main}}\!+\!\frac{1}{\sigma_{2}^{2}} \mathcal{L}_{\mathrm{aux}}\!+\!\log_{}{\sigma_{1}}\!+\!\log_{}{\sigma_{2}}\!+\!\lambda\mathcal{L}_{\mathrm{con}},
    \label{eqn:total_loss}
\end{aligned}
\end{equation} where $\sigma_1$ and $\sigma_2$ represent the uncertainties of MSH and ATHs, which are modeled as learnable parameters. In the early stages of training, the predictions of the ATHs for unlabeled anatomical structures may be inaccurate. Introducing unlabeled data too soon could lead the model to learn misleading information. To address this, we replace $\lambda$ with a ramp-up weighting function $w(t)$, which gradually increases the influence of $\mathcal{L}_{\mathrm{con}}$ as training progresses and mitigates the risk of learning misleading information.

%% file: sec/4_experiments.tex
\begin{figure}[t]
    \centering
    \subfigure[]{
        \begin{minipage}{0.3\linewidth}
            \centering
            \includegraphics[width=\linewidth]{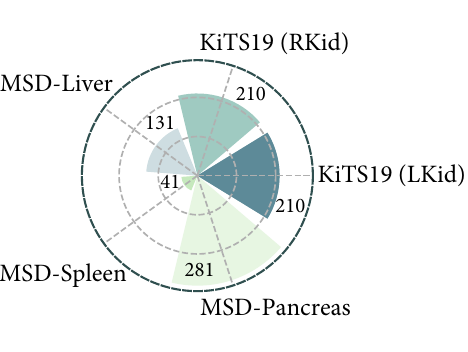}
            \vspace{-0.12in}
            \label{fig:dataset_a}
        \end{minipage}
    }
    \hspace{0.01\linewidth}  
    \subfigure[]{
        \begin{minipage}{0.6\linewidth}
            \centering
            \resizebox{\linewidth}{!}{
                \begin{tabular}{lllll}
                \hline
                    Dataset & Training & Test  & Classes & Annotated Classes \\ \hline
                    KiTS19 & 168   & 42    & 2     & LKid, RKid \\
                    MSD-Liver & 104   & 27    & 1     & Liv \\
                    MSD-Pancreas & 224   & 57    & 1     & Pan \\
                    MSD-Spleen & 32    & 9     & 1     & Spl \\
                    CHAOS-MR & 16    & 4     & 4     & LKid, RKid, Liv, Spl \\ \hline
                    BTCV  &  -     & 30    & 5     & LKid, RKid, Liv, Pan, Spl \\
                    WORD  &  -     & 120   & 5     & LKid, RKid, Liv, Pan, Spl \\
                    AMOS-CT & -    & 300   & 5     & LKid, RKid, Liv, Pan, Spl \\ \hline
                    Total & 544   & 589   & 5     &  LKid, RKid, Liv, Pan, Spl \\
                    \hline
                \end{tabular}
            }
            \vspace{0.1in}
        \end{minipage}
    }
    \vspace{-0.1in}
    \caption{(a): Illustration of the sample size in PLDs. (b): Details of datasets, including dataset splitting, the number of classes, and the annotated classes. The overall classes include left kidney (LKid), right kidney (RKid), liver (Liv), pancreas (Pan), and spleen (Spl).}
    \label{fig:dataset}
    \vspace{-0.2in}
\end{figure}

\section{Experiments}
\subsection{Experimental Setup}
\paragraph{Dataset.}

We perform the experimental evaluation on two benchmarks in the partially labeled setting (PLS), where one is composed of CT data and the other of MR data. For the former, we curate four public CT PLDs—KiTS19~\cite{heller2021state}, MSD-Liver~\cite{antonelli2022medical}, MSD-Pancreas~\cite{antonelli2022medical}, and MSD-Spleen~\cite{antonelli2022medical}—which together comprise a total of 663 volumes. The latter contains 20 MR cases from CHAOS-MR~\cite{kavur2021chaos}. Additionally, we assess the generalization ability of various trained models on three unseen FLDs: BTCV~\cite{landman2015miccai}, WORD~\cite{luo2022word}, and AMOS-CT~\cite{ji2022amos}, containing 450 CT scans. The data split is shown in Fig.~\ref{fig:dataset}, and further dataset details are available in the supplementary material.

\paragraph{Data Preprocessing.}
Following the experimental setup in prior work~\cite{shi2021marginal}, we combine tumor labels with organ labels for the KiTS19, MSD-Liver, and MSD-Pancreas datasets. Additionally, the binary kidney masks in KiTS19 are split into left and right kidneys based on connected components. Furthermore, the original labels of the CHAOS-MR dataset are partitioned into four separate binary categories. To account for the heterogeneity of medical images, a uniform data preprocessing pipeline is applied across all datasets. First, the orientation of all cases is adjusted to the RAS coordinate system. Next, the HU (Hounsfield Unit) values for each sample are truncated to the range [-1024, 1024], and then normalized to the interval [0, 1]. Finally, all data are resampled to a spacing of 1 $\times$ 1 $\times$ 3 $mm^3$.

\paragraph{Implementation Details.}
We use identical training configurations for all models. Each model is trained for 200 epochs with a batch size of 4. During training, we randomly sample patches of size 192 $\times$ 192 $\times$ 64 from each CT scan. The Adam optimizer~\cite{KingBa15} is employed with an initial learning rate of 1e-4. Experiments are implemented with PyTorch on two GeForce RTX 3090 GPUs.

\paragraph{Evaluation Metrics.}
Following prior work~\cite{liu2024cosst}, we evaluate performance using the average Dice Similarity Coefficient (DSC) and the 95th percentile Hausdorff Distance (HD). DSC measures the overlap between the prediction and the ground truth, while HD quantifies the maximum distance between corresponding points in the predicted and ground truth segmentations. A higher DSC and lower HD indicate more accurate segmentation.

\begin{table}
  \centering
  \resizebox{\textwidth}{!}{
  \begin{tabular}{c | cc | cc | cc | cc | cc | cc}
    \toprule
    \multirow{2}[0]{*}{Methods} & \multicolumn{2}{c |}{Left kidney} & \multicolumn{2}{c |}{Right kidney} & \multicolumn{2}{c |}{Liver} & \multicolumn{2}{c |}{Pancreas} & \multicolumn{2}{c |}{Spleen} & \multicolumn{2}{c}{Average} \\ \cline{2-13}
          & DSC $\uparrow$  & HD $\downarrow$    & DSC $\uparrow$  & HD $\downarrow$    & DSC $\uparrow$  & HD $\downarrow$    & DSC $\uparrow$  & HD $\downarrow$    & DSC $\uparrow$  & HD $\downarrow$    & DSC $\uparrow$  & HD $\downarrow$ \\ \hline
    TAL~\cite{fang2020multi}   & \underline{94.57}  & 7.84  & 94.44  & 3.67  & \underline{95.64}  & 5.13  & 80.86  & \textbf{6.24} & 92.95  & \textbf{2.38} & 91.69  & 5.05  \\
    ME~\cite{shi2021marginal}    & 92.04  & 17.08  & 92.57  & 11.13  & 95.21  & 5.04  & 80.15  & 8.62  & 92.40  & 9.07  & 90.47  & 10.19  \\
    DoDNet~\cite{zhang2021dodnet} & 94.29  & 8.16  & \underline{94.67}  & 8.23  & 92.47  & 14.00  & 77.65  & 8.30  & 86.10  & 19.55  & 89.04  & 11.65  \\
    CLIP-Driven~\cite{liu2023clip} & 94.09  & 6.46  & 94.29  & 5.71  & 94.69  & 6.17  & 77.18  & 12.30  & 90.58  & 7.49  & 90.17  & 7.63  \\
    PFL~\cite{liu2022universal}   & 92.85  & 8.91  & 91.48  & 9.38  & 95.23  & 4.78  & 78.12  & 8.82  & 93.20  & 4.99  & 90.17  & 7.38  \\
    Hermes~\cite{gao2024training} & 94.11  & \textbf{4.53}  & 93.63  & 4.29  & 96.05  & \textbf{3.19}  & 59.44  & 13.80  & 89.25  & 10.58  & 86.50  & 7.28  \\
    AAL~\cite{chen2024versatile}   & 94.49  & 5.50  & 94.23  & \underline{3.30}  & 95.59  & 4.36  & \underline{80.89}  & \underline{6.35}  & \textbf{93.83} & \underline{4.34}  & \underline{91.80}  & \textbf{4.77} \\ \hline
    Ours  & \textbf{95.46} & \underline{4.92} & \textbf{94.75} & \textbf{2.49} & \textbf{96.24}  & \underline{3.75}  & \textbf{81.31} & 6.73  & \underline{93.56}  & 6.22  & \textbf{92.26} & \underline{4.82}  \\
    \bottomrule
    \end{tabular}
    }
    \caption{Quantitative comparison with SOTA methods for each anatomical structure on the CT PLDs using average DSC (\%) and HD (95\%). The best and the second results are marked in bold and underlined, respectively.}
  \label{tab:main_result}
\end{table}

\begin{wraptable}{r}{0.5\textwidth}
  \centering
  \caption{Quantitative comparison with SOTA methods on MR PLDs using average DSC (\%).}
  \label{tab:MR}
  \resizebox{\linewidth}{!}{
    \begin{tabular}{c | c | c | c | c | c}
    \toprule
    \multirow{2}[0]{*}{Methods} & \makecell[c]{Left \\ kidney} & \makecell[c]{Right \\ kidney} & Liver  & Spleen & Average \\ \cline{2-6}
          & DSC $\uparrow$  & DSC $\uparrow$   & DSC $\uparrow$  & DSC $\uparrow$  & DSC $\uparrow$ \\ \hline
    TAL~\cite{fang2020multi}        & 82.54  & \textbf{86.90} & \underline{90.33}  & \textbf{85.15} & \underline{86.23}  \\
    ME~\cite{shi2021marginal}       & 82.25  & 76.52  & 86.35  & 84.85  & 82.50  \\
    DoDNet~\cite{zhang2021dodnet}   & 83.12  & 72.24  & 86.14  & 79.51  & 80.25  \\
    CLIP-Driven~\cite{liu2023clip}  & 81.35  & 80.87  & 87.61  & 84.98  & 83.70  \\
    PFL~\cite{liu2022universal}     & 80.12  & 79.59  & 88.61  & 83.26  & 82.90  \\
    Hermes~\cite{gao2024training}   & 61.98  & 63.98  & 89.92  & 71.42  & 71.82  \\
    AAL~\cite{chen2024versatile}    & \underline{85.32}  & 84.84  & 84.26  & 84.03  & 84.61  \\ \hline
    Ours        & \textbf{87.20} & \underline{85.50}  & \textbf{91.36} & \underline{84.85}  & \textbf{87.23} \\
    \bottomrule
    \end{tabular}%
  }
\end{wraptable}

\subsection{Results on Partially Labeled Datasets}

\begin{figure}
\centerline{\includegraphics[width=1.0\textwidth]{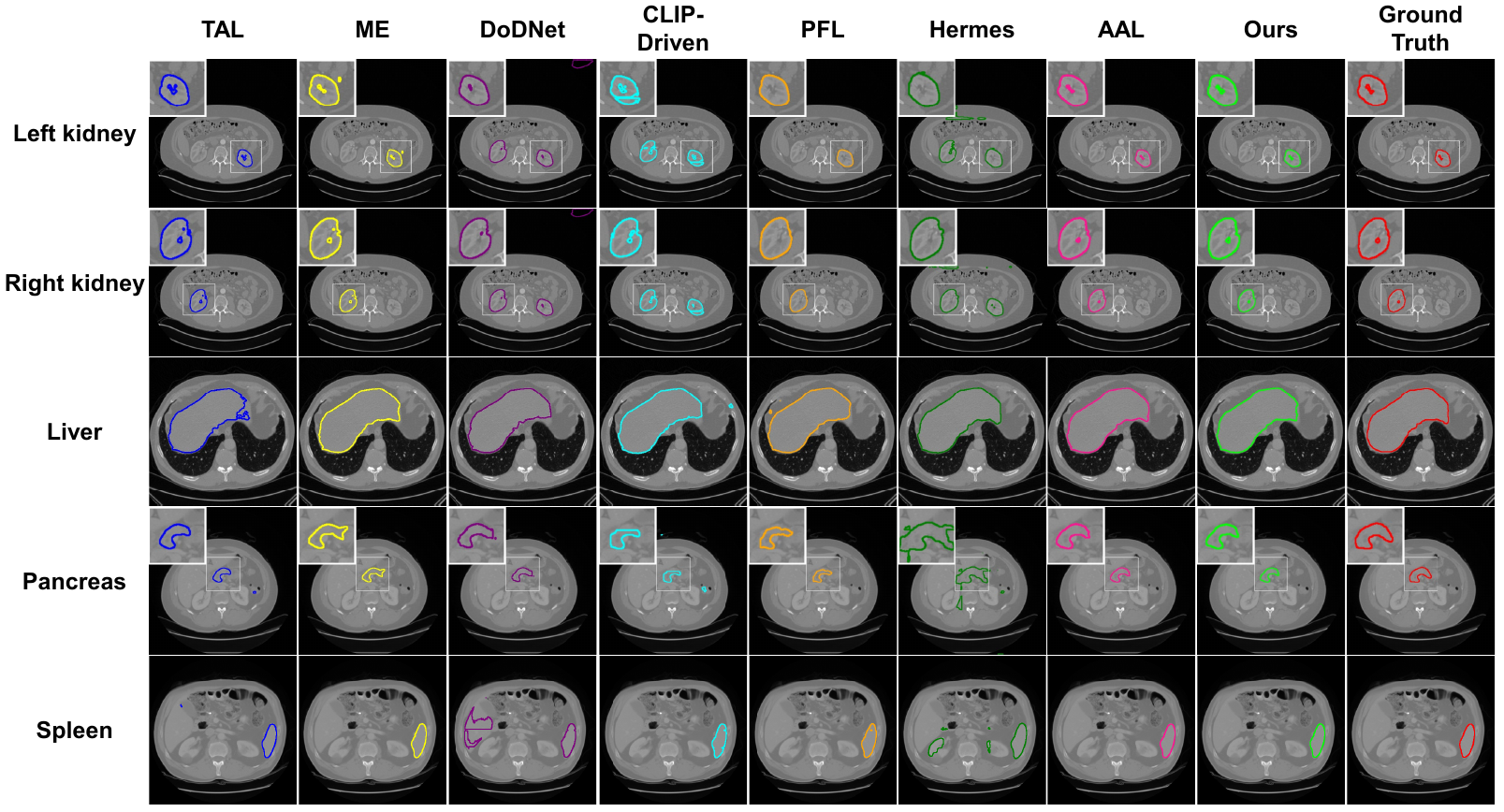}}
\vspace{-0.1in}
\caption{Visual comparison of segmentation results between our proposed and other methods of comparison on the CT PLDs.}
\label{fig:visualization}
\end{figure}

\paragraph{Quantitative Results.}
As shown in Tab.~\ref{tab:main_result}, we evaluate our method on the CT PLDs. First, loss optimization methods (TAL and ME) outperform task prompt strategies (DoDNet, CLIP-Driven, and Hermes) in average performance, due to the inherent anatomical exclusivity of the former, which the latter lacks (see Fig.~\ref{fig:visualization}). Second, the pseudo-full label method (PFL) suffers from noisy pseudo labels, leading to reduced segmentation accuracy. Third, our method achieves the highest DSC values in most tasks, except for the Spleen task, where it slightly underperforms compared to AAL. Overall, the results demonstrate that our approach outperforms all compared methods, achieving an average DSC of 92.26\% and HD of 4.82. The MR results in Tab.~\ref{tab:MR} further demonstrate that our method maintains its leading position, achieving an average DSC of 87.23\%.

\paragraph{Qualitative Results.}
To further demonstrate the superiority of our method, we visualize the segmentation results of different approaches on the CT PLDs in Fig.~\ref{fig:visualization}. An interesting observation arises during the segmentation of the left and right kidneys: task prompt models (DoDNet, CLIP-Driven, and Hermes) tend to simultaneously segment the contralateral kidney (\textit{e.g.}, the right kidney for the left kidney, and vice versa). This occurs because the predictions of each task in these models are independent, lacking anatomical structure exclusivity. In contrast, our method distinctly identifies and separates the left and right kidneys. Overall, the visualization highlights that our method achieves more accurate segmentation compared to the other approaches.

\begin{figure}[t]
    \centering
    \subfigure[]{           
        \includegraphics[width=0.32\linewidth]{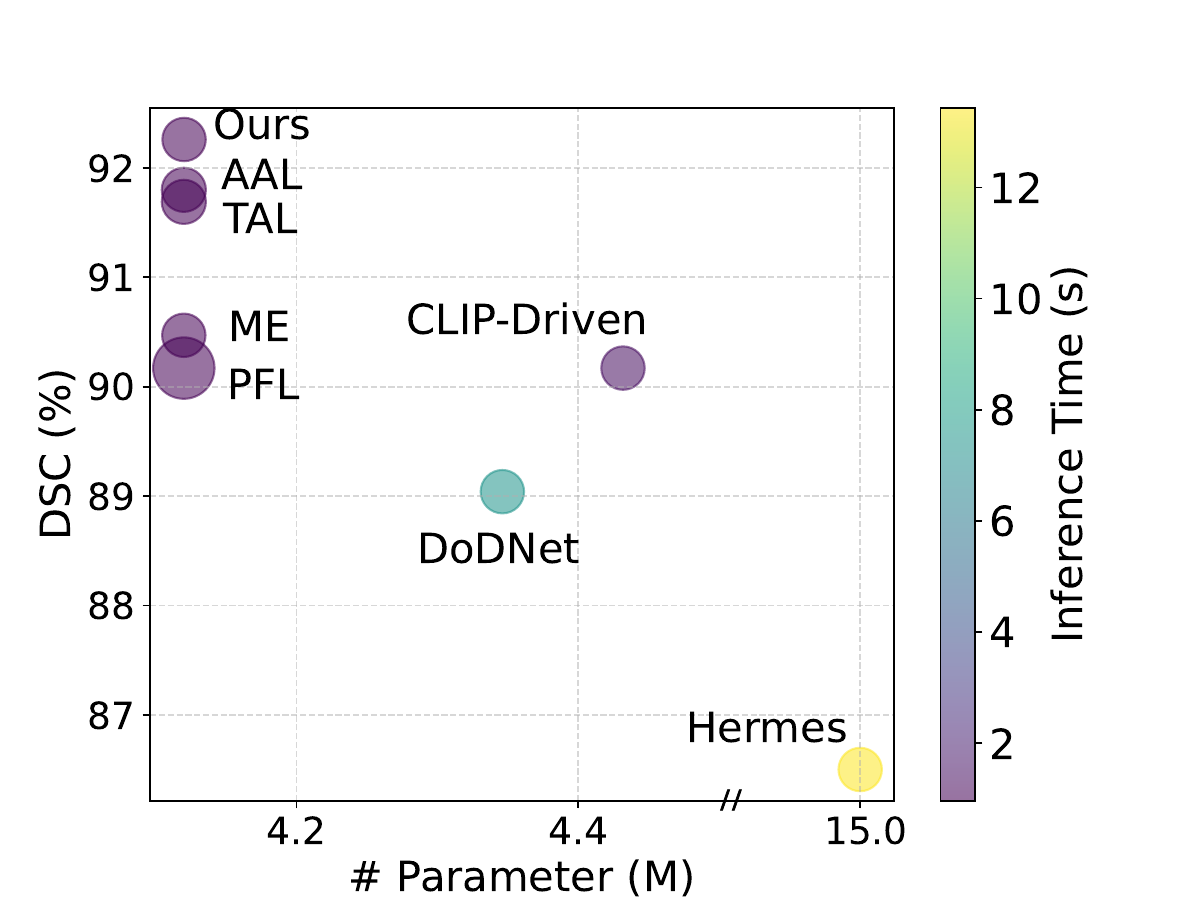} 
        \label{fig:bubble_graph}  
    }
    \subfigure[]{           
        \includegraphics[width=0.31\linewidth]{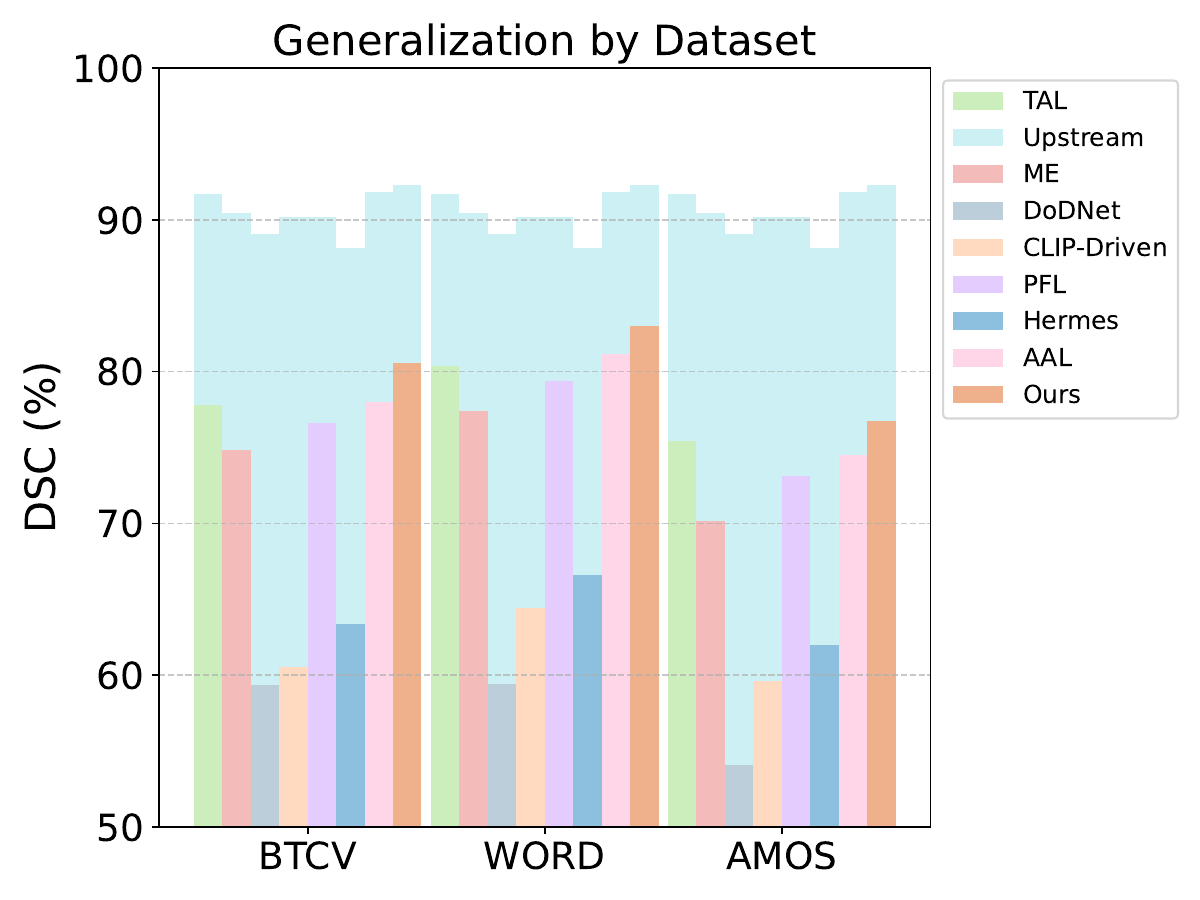} 
        \label{fig:generalization_dataset}         
    }
    \subfigure[]{           
        \includegraphics[width=0.31\linewidth]{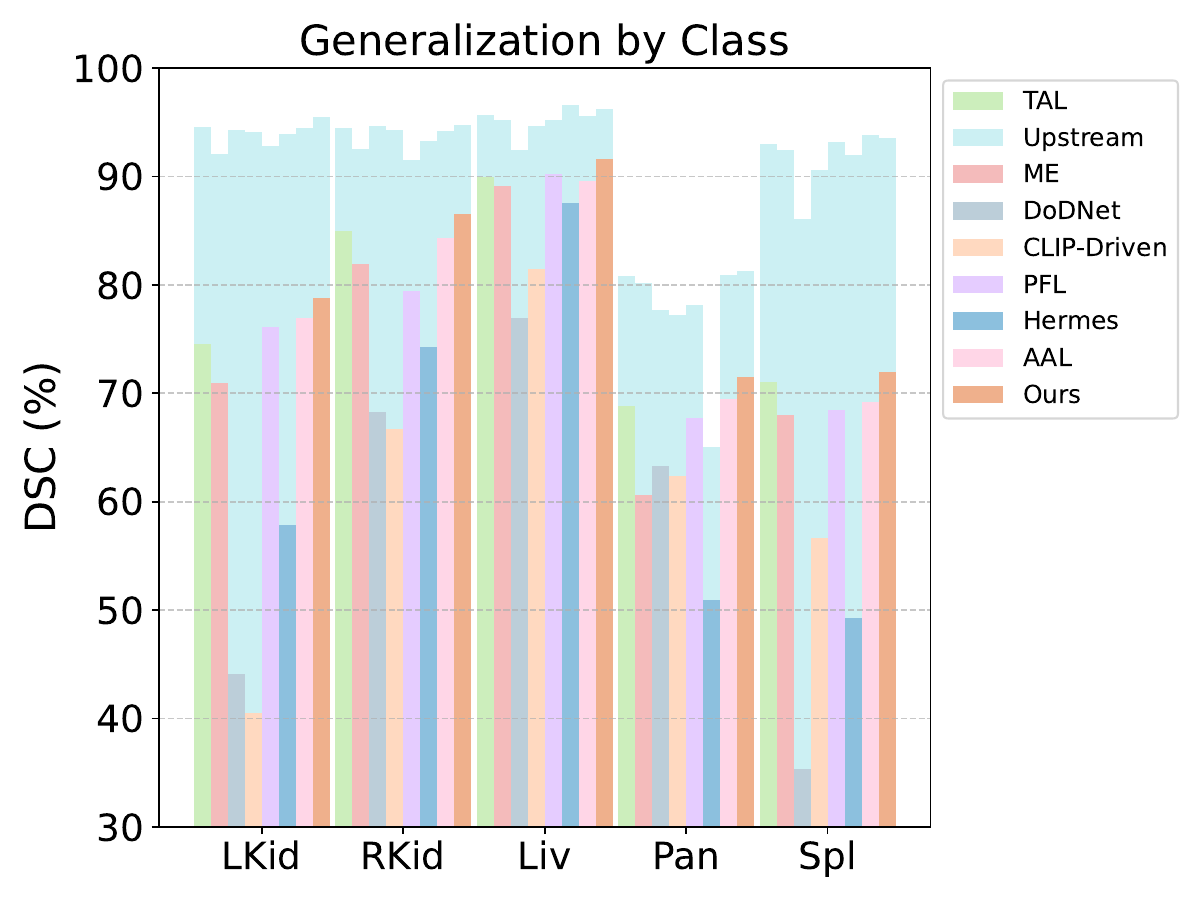} 
        \label{fig:generalization_task}    
    }
    \vspace{-0.1in}
    \caption{ (a) Comparison with other methods in terms of parameter number, average DSC scores, training time, and inference time on the CT PLDs. For each data point, the size of the bubble is proportional to the training time. A larger bubble corresponds to a longer training time, and vice versa. The inference time is computed with an input size of 192 $\times$ 192 $\times$ 64. `M' denotes million. (b) Generalization comparison of different methods from PLDs to FLDs. The statistics are calculated as the average DSC values across all classes for each dataset. `Upstream' represents the results on the PLDs. (c) Generalization comparison of different methods from PLDs to FLDs. The statistics are calculated based on the mean DSC values for identical classes across all datasets. }
    \label{fig:three_graphs}
    \vspace{-0.2in}
\end{figure}

\begin{wraptable}{r}{0.6\textwidth}
  \centering
  \caption{Segmentation performance comparison with SOTA methods on the CT PLDs in a low-data regime, in terms of average DSC (\%) and HD (95\%) across five tasks.}
\label{tab:portion}
  \resizebox{\linewidth}{!}{
    \begin{tabular}{c | cc | cc | cc}
    \toprule
    \multirow{2}[0]{*}{Methods} & \multicolumn{2}{c |}{20\%} & \multicolumn{2}{c |}{50\%} & \multicolumn{2}{c}{100\%} \\ \cline{2-7}
          & DSC $\uparrow$  & HD $\downarrow$ & DSC $\uparrow$ & HD $\downarrow$ & DSC $\uparrow$ & HD $\downarrow$ \\ \hline
    TAL~\cite{fang2020multi}        & 84.10  & 20.33  & 89.53  & 7.60  & 91.69  & 5.05  \\
    ME~\cite{shi2021marginal}       & 84.50  & 12.80  & 88.83  & 9.67  & 90.47  & 10.19  \\
    DoDNet~\cite{zhang2021dodnet}   & 85.09  & 16.46  & 88.43  & 12.88  & 89.04  & 11.65  \\
    CLIP-Driven~\cite{liu2023clip}  & 80.74  & 23.74  & 81.55  & 19.58  & 90.17  & 7.63  \\
    PFL~\cite{liu2022universal}     & 87.17  & 12.88  & 90.13  & 6.88  & 90.17  & 7.38  \\
    Hermes~\cite{gao2024training}   & 80.75  & 16.62  & 84.22  & 14.86  & 86.50  & 7.28  \\
    AAL~\cite{chen2024versatile}    & 87.01  & 11.25  & 90.69  & 6.02  & 91.80  & \textbf{4.77} \\ \hline
    Ours  & \textbf{87.84} & \textbf{7.94} & \textbf{90.94} & \textbf{5.83} & \textbf{92.26} & 4.82 \\
    \bottomrule
    \end{tabular}%
  }
\end{wraptable}

\paragraph{Comparison in Low-Data Regime.}
We further investigate the impact of training data size on model performance with CT PLDs by training different models on random subsets of the data, using 20\% and 50\% portions. The average performance across five tasks is presented in Tab.~\ref{tab:portion}. The results show that our method outperforms the other models in low-data regimes, demonstrating the effectiveness of task consistency training in leveraging valuable unlabeled information to enhance segmentation performance when data is limited.

\paragraph{Model Efficiency.}

We compare the number of parameters, average DSC scores, inference time, and training time among all methods in Fig.~\ref{fig:bubble_graph}. Several models, including ours, AAL~\cite{chen2024versatile}, TAL~\cite{fang2020multi}, ME~\cite{shi2021marginal}, and PFL~\cite{liu2022universal}, have the minimal parameter count of 4.12M. Note that the ATHs in our method are not used during inference. In contrast, DoDNet and CLIP-Driven require additional controllers for task prompts, increasing their parameter counts to 4.3M and 4.4M, respectively. Hermes, which uses learnable prior vectors across multiple stages of the encoder and decoder, has a significantly higher parameter count of 14.9M.
Regarding inference time, DoDNet and Hermes require multiple inferences to obtain predictions for all targets, leading to longer inference times that increase with the number of targets. Other methods, including ours, exhibit similar inference times, approximately one-fifth that of DoDNet. Additionally, PFL requires training five individual networks to generate pseudo-fully labeled datasets, resulting in longer training times compared to other methods. In summary, our approach outperforms the others in segmentation accuracy, training time, inference speed, and model size. Further details on model efficiency comparisons are available in the supplementary material.

\subsection{Results on Fully Labeled Datasets}

\paragraph{Generalization.}
To evaluate the generalization ability of our model, we perform inference on unseen FLDs using models trained on partially labeled datasets. The evaluation includes the BTCV (30 samples), WORD (120 samples), and AMOS-CT (300 samples) datasets. The results, shown in Fig.~\ref{fig:generalization_dataset} and Fig.~\ref{fig:generalization_task}, reveal that all models experience a performance decline on FLDs compared to PLDs (denoted as `Upstream'). Notably, DoDNet and CLIP-Driven show a more significant drop in performance. In contrast, our method consistently outperforms others across all datasets and classes, highlighting its superior generalization ability. This improvement is attributed to the TCT framework, which enables the model to effectively leverage unlabeled anatomical structures. Detailed quantitative results can be found in the supplementary material.

\begin{wraptable}{r}{0.5\textwidth}
  \centering
  \caption{Comparison of transfer learning segmentation results among different methods on the WORD dataset using average DSC (\%) and HD (95\%) across five tasks.}
\label{tab:transfer}
  \resizebox{\linewidth}{!}{
    \begin{tabular}{c | cc | cc | cc}
    \toprule
    \multirow{2}[0]{*}{Methods} & \multicolumn{2}{c |}{20\%} & \multicolumn{2}{c |}{50\%} & \multicolumn{2}{c}{100\%} \\ \cline{2-7}
          & DSC $\uparrow$  & HD $\downarrow$ & DSC $\uparrow$ & HD $\downarrow$ & DSC $\uparrow$ & HD $\downarrow$ \\ \hline
    From Scratch   & 79.33  & 17.15  & 86.60  & 5.25  & 89.11  & 5.42  \\ \hline
    TAL~\cite{fang2020multi}   & 86.96  & 7.44  & 87.38  & 5.63  & 89.63  & 3.95  \\
    ME~\cite{shi2021marginal}    & 86.37  & 6.65  & 87.69  & 5.55  & 89.36  & 5.09  \\
    DoDNet~\cite{zhang2021dodnet} & 60.59  & 59.54  & 69.84  & 50.93  & 71.08  & 49.57  \\
    CLIP-Driven~\cite{liu2023clip} & 54.23  & 49.32  & 57.37  & 51.18  & 60.69  & 39.00  \\
    PFL~\cite{liu2022universal}   & 85.94  & 7.87  & 87.54  & 6.63  & 89.60  & \textbf{3.27} \\
    Hermes~\cite{gao2024training} & 67.89  & 33.55  & 71.83  & 24.70  & 72.83  & 18.34  \\
    AAL~\cite{chen2024versatile}   & 86.28  & 7.14  & 87.48  & 5.13  & 89.59  & 3.98  \\ \hline
    Ours  & \textbf{87.37}  & \textbf{4.92} & \textbf{88.27}  & \textbf{4.65} & \textbf{89.77}  & 3.47  \\
    \bottomrule
    \end{tabular}%
  }
\end{wraptable}

\paragraph{Transfer Learning.}
Given the challenge of obtaining large-scale fully annotated datasets, we fine-tune a model pre-trained on partially labeled datasets for downstream tasks. To assess segmentation performance under low-data conditions, we use the WORD~\cite{luo2022word} dataset, splitting it into a training set of 96 samples and a testing set of 24 samples. The model is fine-tuned using random subsets of the training data at 20\%, 50\%, and 100\%. The average DSC (\%) and HD (95\%) scores for the five tasks are reported in Tab.~\ref{tab:transfer}. Our results show that most pre-trained models outperform those trained from scratch, except for DoDNet and CLIP-Driven models. By comparing different experimental setups, we can observe that our method consistently leads in segmentation performance, demonstrating that leveraging unlabeled data in upstream tasks can improve segmentation accuracy in downstream tasks.

\subsection{Ablation Studies}

\begin{wraptable}{r}{0.5\textwidth}
  \centering
  \caption{Performance comparison for different components on the partially labeled test set using average DSC (\%) and HD (95\%).}
  \label{tab:ablation}
  \resizebox{\linewidth}{!}{
    \begin{tabular}{p{1.8cm}<{\centering} | p{1.8cm}<{\centering} | p{1.8cm}<{\centering} | p{1.8cm}<{\centering} | p{1.8cm}<{\centering}}
    \toprule
    \makecell[c]{TCT} & \makecell[c]{Consistency \\ Filtering} & \makecell[c]{UAUWL} & DSC $\uparrow$ & HD $\downarrow$ \\ \midrule
    \xmark       & \xmark & \xmark & 91.67  & 6.40  \\
    \cmark       & \xmark & \xmark & 91.86  & 5.04  \\
    \cmark       & \cmark & \xmark & 92.10  & 4.98  \\
    \cmark       & \xmark & \cmark & 92.03  & 5.33  \\
    \cmark       & \cmark & \cmark & \textbf{92.30} & \textbf{4.71}  \\
    \bottomrule
    \end{tabular}
  }
\end{wraptable}

\paragraph{Effectiveness of Different Components.}
To evaluate the effectiveness of the core components, we conduct ablation studies on the partially labeled datasets. As shown in Tab.~\ref{tab:ablation}, we report the average DSC (\%) and HD (95\%) for the five tasks. The results demonstrate that the TCT framework improves the DSC from 91.67\% to 91.86\%, highlighting its ability to leverage unlabeled structures. Incorporating consistency filtering to exclude potentially noisy low-consistency data further increases the DSC to 92.10\%. Adding UAUWL with TCT alone yields a DSC of 92.03\%, suggesting that UAUWL effectively balances the MSH and ATHs. Finally, the combination of all three components achieves the highest performance, with an average DSC of 92.30\% and the lowest HD of 4.71, indicating the complementary benefits of these components in improving both segmentation accuracy and boundary precision. Additional ablation studies (\textit{e.g.}, threshold selection) are provided in the supplementary.

%% file: sec/5_conclusion.tex
\section{Conclusion}

In this paper, we propose a task consistency training framework for versatile medical image segmentation from partially labeled datasets. By leveraging unlabeled structures in each partially labeled example, TCT alleviates the class imbalance issue without requiring additional training processes. We further introduce a consistency filtering strategy to exclude potentially noisy low-consistency data, preventing error accumulation and propagation. Additionally, we present a unified auxiliary uncertainty-weighted loss to avoid segmentation performance degradation caused by the dominance of specific tasks. Through experiments on low-data regimes and model efficiency, we highlight the advantages of our model from multiple perspectives. Moreover, we further investigate its generalization capability and applicability in the settings of fully labeled datasets and transfer learning. Overall, extensive experiments across eight public datasets with 1,133 cases demonstrate the superiority of our method over state-of-the-art versatile segmentation models.

%% file: sec/X_suppl.tex
\section{Discussion}

\paragraph{Rationale of TCT.}
Class imbalance in PLDs stems from unequal category distributions, resulting in label scarcity for certain classes. ATHs generate task-specific predictions as pseudo-labels without requiring additional training processes, guiding MSH learning via a task consistency constraint. Additionally, our consistency filtering strategy removes potentially noisy pseudo-labels, enabling the integration of information from unlabeled classes and mitigating class imbalance effectively.

\paragraph{Limitations and Future Work.}
While our proposed method leverages different ATHs to generate task-specific predictions to enhance the learning of the MSH, it does not consider the anatomical similarities observed among some organs. We suggest that associations may exist between certain ATHs, such as the left and right kidneys, and explicitly modeling these correlations could enhance segmentation performance in the future. We hope this work inspires future research to tackle more challenging versatile segmentation scenarios.

\section{More Details about Network Architecture}

The configuration of the 3D U-Net~\cite{cciccek20163d} architecture is depicted in Tab.~\ref{tab:network_architecture}. 3D U-Net is a network architecture composed of symmetrical encoder and decoder structures. It consists of five stages, with each stage containing two 3 $\times$ 3 $\times$ 3 convolutional layers, except for the last one. As shown in Tab.~\ref{tab:network_architecture}, each number within the curly braces indicates the input or output channels of one convolutional layer. Skip connections~\cite{he2016deep} link the encoder and decoder, preserving detailed low-level information. Connections between adjacent stages are made through downsampling and upsampling, where downsampling is achieved via 2 $\times$ 2 $\times$ 2 max pooling, and upsampling is implemented using trilinear interpolation. For the MSH and each ATH, we utilize a 3 $\times$ 3 $\times$ 3 convolution layer to produce segmentation results.

\begin{table}[h]
  \caption{Details of network architecture. Each pair of curly braces represents a convolutional layer, with the numbers inside indicating the input and output channels.}
        \label{tab:network_architecture}
  \centering
 \begin{tabular}{l | l | l}
            \toprule
             & Encoder & Decoder \\
            \midrule
            \multirow{2}{*}{Stage 1}    & \{1, 16\} & \{48, 16\} \\
                                        & \{16, 16\} & \{16, 16\} \\
            \midrule
            \multirow{2}{*}{Stage 2}    & \{16, 16\} & \{96, 32\} \\
                                        & \{16, 32\} & \{32, 32\} \\
            \midrule
            \multirow{2}{*}{Stage 3}    & \{32, 32\} & \{192, 64\} \\
                                        & \{32, 64\} & \{64, 64\} \\
            \midrule
            \multirow{2}{*}{Stage 4}    & \{64, 64\} & \{384, 128\} \\
                                        & \{64, 128\} & \{128, 128\} \\
            \midrule
            {Stage 5} & \{128, 128\} & \{128, 256\} \\
            \bottomrule
         \end{tabular}
\end{table}

\section{More Details about Datasets}
We introduce the details of the datasets as follows, including the dataset sizes, modalities, and anatomical annotations.

\begin{itemize}[leftmargin=*,noitemsep,topsep=0pt]
    \item \textbf{KiTS19:} The KiTS19~\cite{heller2021state} dataset comprises 210 abdominal CT volumes with manually delineated kidney and tumor labels.

    \item \textbf{MSD-Liver \& Pancreas \& Spleen:} These three datasets are part of the MSD~\cite{antonelli2022medical} collection. MSD-Liver contains 3D CT scans from 131 patients, primarily used for liver and liver tumor segmentation tasks. MSD-Pancreas focuses on the task of the pancreas and pancreatic tumor segmentation, including a large number of CT scans from 281 patients. MSD-Spleen consists of 41 CT images with spleen organ annotation.

    \item \textbf{CHAOS-MR:} CHAOS-MR~\cite{kavur2021chaos} is a widely used benchmark for assessing MRI-based multi-organ segmentation performance. It consists of 20 cases, each comprising multiple MRI slices covering the abdominal region.

    \item \textbf{BTCV:} The BTCV~\cite{landman2015miccai} dataset is a key benchmark in the field of abdominal multi-organ segmentation. It comprises 30 abdominal CT scan sequences, each annotated with 13 different organ labels.

    \item \textbf{WORD:} The WORD~\cite{luo2022word} is a comprehensive collection of abdominal CT scans. The dataset includes 120 CT scans fully covering the abdominal area, with detailed labels for 16 different abdominal organs.

    \item \textbf{AMOS-CT:} AMOS-CT~\cite{ji2022amos} is a medical imaging dataset specifically designed for evaluating and developing algorithms for multi-organ segmentation in abdominal CT images. AMOS-CT contains 300 samples, providing manual annotations for 15 key abdominal organs.
    
\end{itemize}

\section{More Details about Baselines}
We compare our method with seven recent versatile segmentation models that learn from partially labeled datasets. To ensure fairness, we use 3D-UNet~\cite{cciccek20163d} as the backbone for all compared models. The methods included in the comparison are as follows:

\begin{itemize}[leftmargin=*,noitemsep,topsep=0pt]
    \item \textbf{TAL}~\cite{fang2020multi}
    is trained on the partially labeled ground truth, treating unlabeled pixels as background. TAL produces multi-channel predictions for all tasks.

    \item \textbf{ME}~\cite{shi2021marginal}
    applies an exclusive constraint to the unlabeled targets, building on the foundation provided by TAL. Like TAL, ME generates multi-channel predictions.

    \item \textbf{DoDNet}~\cite{zhang2021dodnet}
    trains with partial labels and outputs single-channel predictions for each target by incorporating one-hot task priors.

    \item \textbf{CLIP-Driven}~\cite{liu2023clip}
    is trained on PLDs and generates multi-channel predictions using CLIP embeddings.

    \item \textbf{PFL}~\cite{liu2022universal}
    trains with pseudo-full labels and generates multi-channel predictions.

    \item \textbf{Hermes}~\cite{gao2024training}
    incorporates learnable task and modality priors at multiple stages of the network to perform various segmentation tasks, similar to DoDNet~\cite{zhang2021dodnet}.

    \item \textbf{AAL}~\cite{chen2024versatile}
    uses an ambiguity-aware loss to address label ambiguity in partial labels.
    
\end{itemize}

\section{More Experimental Results}

\subsection{More Ablation Studies}

\paragraph{Impact of the Threshold for Consistency Filtering.}
We conduct experiments to explore the impact of the threshold $\theta$ for the consistency filtering strategy, including fixed value (0.5), the mean IoU of the batch for each ATH (Batch Mean), the mean IoU of all ATHs (Task Mean), and the median IoU of all ATHs (Task Median). As shown in Tab.~\ref{tab:theta}, the Batch Mean only reaches 77.66\% of DSC, potentially affected by the batch size. In contrast, a fixed value (0.5) and Task Mean eliminate the effect of batch size, increasing DSC to 86.02\% and 85.62\%, respectively. Moreover, we find Task Median achieves the highest DSC of 87.55\% and the lowest HD of 7.04\%. This can be attributed to the robustness of the median in handling the outliers. Thus we use the Task Median as the threshold $\theta$ for all experiments.

\begin{table}[h]
  \caption{Performance comparison across different thresholds ($\theta$) using average DSC (\%) and HD (95\%). CF represents consistency filtering.}
\label{tab:theta}
  \centering
  \begin{tabular}{r | c | c | c | c | c}
    \toprule
    Metric & w/o CF & 0.5  & Batch Mean & Task Mean & Task Median \\ \midrule
    DSC $\uparrow$  & 87.34 & 86.02  & 77.66 & 85.62 & \textbf{87.55} \\
    HD $\downarrow$ & 9.73 & 9.39  & 28.80 & 9.06 & \textbf{7.40}  \\
    \bottomrule
    \end{tabular}%
\end{table}

\begin{table}[h]
  \caption{Performance comparison of IoU filtering and confidence thresholding strategies for consistency filtering.}
\label{tab:filter_strategy}
  \centering
  \begin{tabular}{c | c | c }
    \toprule
    \makecell[c]{Consistency \\ Filtering Strategy} & DSC $\uparrow$ & HD $\downarrow$  \\ \midrule
    Confidence  & 77.23 & 13.47 \\
    IoU  & \textbf{87.55}  & \textbf{7.40}  \\
    \bottomrule
    \end{tabular}%
\end{table}

\begin{table}[h]
  \caption{Performance comparison between our proposed UAUWL and the previous UWL~\cite{kendall2018multi, xu2022mtformer}.}
\label{tab:uncertainty}
  \centering
  \begin{tabular}{c | c | c }
    \toprule
    \makecell[c]{Uncertainty \\ Loss} & DSC $\uparrow$ & HD $\downarrow$  \\ \midrule
    UWL  & 87.00 & 9.60 \\
    UAUWL & \textbf{87.39}  & \textbf{8.08}  \\
    \bottomrule
    \end{tabular}%
\end{table}

\paragraph{Impact of Consistency Filtering Strategies.}
To investigate the impact of consistency filtering strategies on model performance, we compare the IoU filtering with confidence thresholding, as shown in Tab.~\ref{tab:filter_strategy}. To implement confidence thresholding, we quantify the confidence of ATHs using entropy and discard predictions with confidence below 0.5. As shown in the results, IoU filtering significantly outperforms confidence thresholding, achieving a much higher DSC of 87.55 versus 77.23 and a lower HD of 7.40 compared to 13.47. We therefore adopt the IoU metric as the consistency filtering strategy.

\paragraph{Effectiveness of UAUWL.}
To assess the effectiveness of the proposed UAUWL, we compare it with the baseline UWL~\cite{kendall2018multi, xu2022mtformer} in terms of DSC and HD, as summarized in Tab.~\ref{tab:uncertainty}. Our method achieves a higher DSC of 87.39 compared to 87.00 for UWL, indicating improved segmentation accuracy. In addition, UAUWL yields a substantially lower HD of 8.08 versus 9.60 from UWL, suggesting better boundary precision. These results demonstrate that UAUWL effectively enhances both the overlap and the contour quality of segmentation, outperforming the previous UWL. As a side note, all methods use 3D-UNet as the backbone when reporting the number of parameters.

\subsection{Detailed Comparison of Model Efficiency}
In Tab.~\ref{tab:detailed_bubble}, we show a detailed comparison of model efficiency on the partially labeled datasets. Note that we calculate the parameter number of the model during inference. The training time records the duration of each epoch of the model. These findings are consistent with those shown in Fig.~\ref{fig:bubble_graph}, confirming that our model excels over the compared models regarding segmentation accuracy, number of parameters, training duration, and inference speed.

\begin{table}[h]
  \caption{Detailed comparison results on the partially labeled datasets in terms of parameter number, training time, inference time, and average DSC scores.}
\label{tab:detailed_bubble}
  \centering
  \begin{tabular}{l | l | l | l | l}
    \toprule
    Methods & \#Params (M) & \makecell[l]{Training \\ Time (min.s)} & \makecell[l]{Inference \\ Time (s)} & DSC (\%) \\ \midrule
    TAL~\cite{fang2020multi}   & 4.12 & 29.58 & 0.96  & 91.69 \\
    ME~\cite{shi2021marginal}    & 4.12 & 29.46 & 0.96  & 90.47 \\
    DoDNet~\cite{zhang2021dodnet} & 4.34 & 29.02 & 7.43  & 89.04 \\
    CLIP-Driven~\cite{liu2023clip} & 4.43 & 29.07 & 1.39  & 90.17 \\
    PFL~\cite{liu2022universal}   & 4.12 & 58.47 & 0.96  & 90.17 \\
    Hermes~\cite{gao2024training} & 14.90 & 29.39 & 13.43  & 86.50  \\
    AAL~\cite{chen2024versatile}   & 4.12 & 30.11 & 0.96  & 91.80  \\ \midrule
    Ours  & 4.12 & 29.49 & 0.96  & 92.26  \\
    \bottomrule
    \end{tabular}%
\end{table}

\begin{figure}[h]
\centerline{\includegraphics[width=1.0\textwidth]{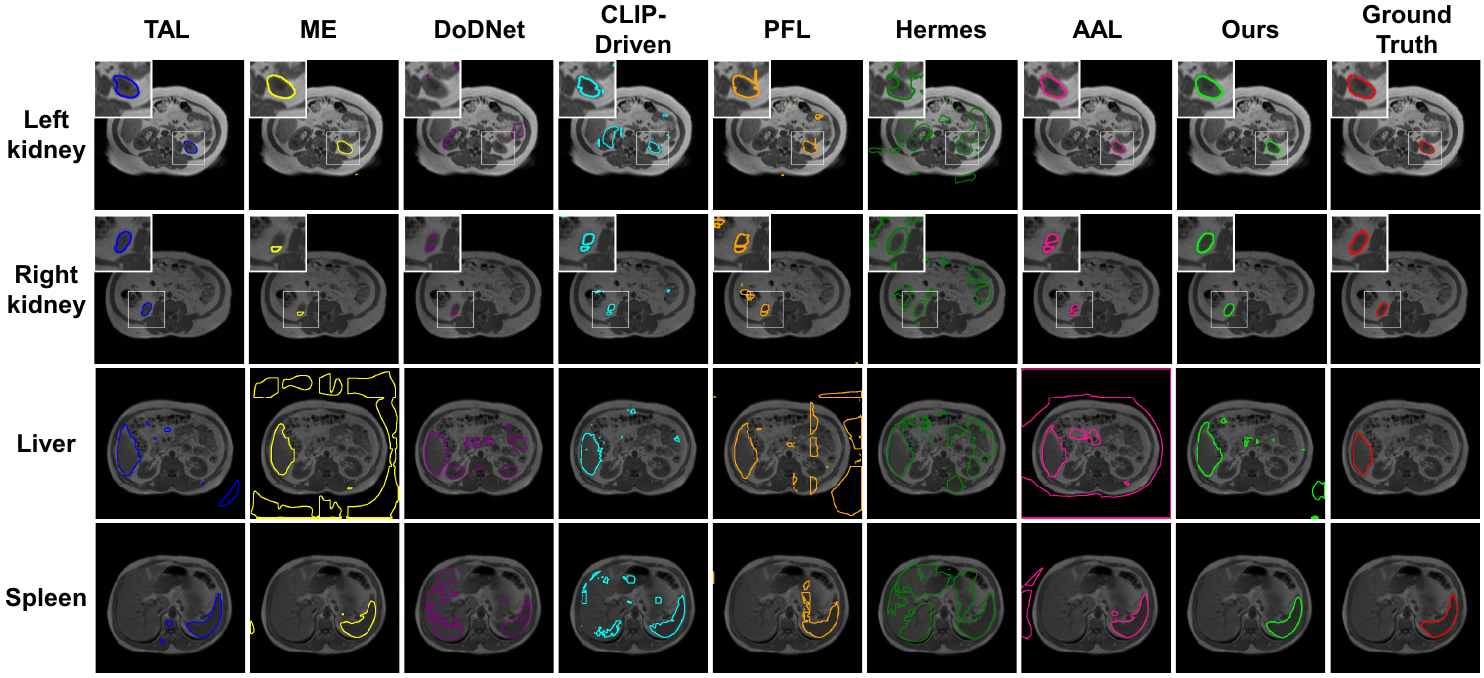}}
\caption{Visual comparison of segmentation results between our proposed and other methods of comparison on the MR PLDs.}
\label{fig:visualization_new}
\end{figure}

\subsection{Qualitative Results on CHAOS-MR Dataset}
We show the visualization results on CHAOS-MR in Fig.~\ref{fig:visualization_new}, a qualitative version of Tab.~\ref{tab:MR}, which further prove the superiority of our proposed method for achieving accurate segmentation in the partially labeled setting.

\subsection{Detailed Generalization Comparison on Fully Labeled Datasets}
We present detailed generalization comparison results for each dataset and anatomical structure in Tab.~\ref{tab:detailed_by_dataset} and Tab.~\ref{tab:detailed_by_organ}, respectively. These results represent the detailed numerical versions of Fig.~\ref{fig:generalization_dataset} and Fig.~\ref{fig:generalization_task}. It can be observed that our method maintains a leading position across various datasets and anatomical structures, demonstrating its strong generalization capability.

\begin{table}[h]
  \caption{The Detailed generalization comparison results of Fig.~\ref{fig:generalization_dataset} on each dataset using average DSC (\%) and HD (95\%).}
\label{tab:detailed_by_dataset}
  \centering
  \begin{tabular}{c | cc | cc | cc}
    \toprule
    \multirow{2}[0]{*}{Methods} & \multicolumn{2}{c |}{BTCV} & \multicolumn{2}{c |}{WORD} & \multicolumn{2}{c}{AMOS} \\ \cline{2-7}
          & DSC $\uparrow$  & HD $\downarrow$ & DSC $\uparrow$ & HD $\downarrow$ & DSC $\uparrow$ & HD $\downarrow$ \\ \hline
    TAL~\cite{fang2020multi}        & 77.83  & 18.87  & 80.35  & 16.57  & 75.41  & 18.38  \\
    ME~\cite{shi2021marginal}       & 74.82  & 28.06  & 77.38  & 22.96  & 70.17  & 26.97  \\
    DoDNet~\cite{zhang2021dodnet}   & 59.32  & 57.60  & 59.43  & 55.94  & 54.06  & 59.88  \\
    CLIP-Driven~\cite{liu2023clip}  & 60.53  & 54.98  & 64.43  & 50.00  & 59.61  & 55.30  \\
    PFL~\cite{liu2022universal}     & 76.58  & 16.49  & 79.41  & 12.03  & 73.14  & 16.22  \\
    Hermes~\cite{gao2024training}   & 63.34  & 45.94  & 66.59  & 42.33  & 61.98  & 49.39  \\
    AAL~\cite{chen2024versatile}    & 77.97  & 17.11  & 81.19  & 13.40  & 74.51  & 16.56  \\ \hline
    Ours        & \textbf{80.55} & \textbf{11.94} & \textbf{82.97} & \textbf{10.05} & \textbf{76.77} & \textbf{12.11} \\
    \bottomrule
    \end{tabular}%
\end{table}

\begin{table}[h]
\caption{Detailed generalization comparison of Fig.~\ref{fig:generalization_task} on each class using average DSC (\%) and HD (95\%)}
  \label{tab:detailed_by_organ}
  \centering
  \resizebox{\textwidth}{!}{
  \begin{tabular}{c | cc | cc | cc | cc | cc | cc}
    \toprule
    \multirow{2}[0]{*}{Methods} & \multicolumn{2}{c |}{Left kidney} & \multicolumn{2}{c |}{Right kidney} & \multicolumn{2}{c |}{Liver} & \multicolumn{2}{c |}{Pancreas} & \multicolumn{2}{c |}{Spleen} & \multicolumn{2}{c}{Average} \\ \cline{2-13}
          & DSC $\uparrow$  & HD $\downarrow$    & DSC $\uparrow$  & HD $\downarrow$    & DSC $\uparrow$  & HD $\downarrow$    & DSC $\uparrow$  & HD $\downarrow$    & DSC $\uparrow$  & HD $\downarrow$    & DSC $\uparrow$  & HD $\downarrow$ \\ \hline
    TAL~\cite{fang2020multi}   & 74.50  & 24.27  & 85.00  & 9.93  & 89.97  & 10.05  & 68.78  & 13.25  & 71.07  & 32.20  & 77.86  & 17.94  \\
    ME~\cite{shi2021marginal}    & 70.90  & 35.77  & 81.92  & 25.12  & 89.14  & 10.67  & 60.65  & 22.85  & 68.01  & 35.59  & 74.12  & 26.00  \\
    DoDNet~\cite{zhang2021dodnet} & 44.14  & 76.97  & 68.29  & 48.23  & 76.94  & 32.38  & 63.28  & 25.75  & 35.38  & 105.70  & 57.60  & 57.81  \\
    CLIP-Driven~\cite{liu2023clip} & 40.51  & 76.81  & 66.66  & 45.04  & 81.45  & 28.80  & 62.34  & 37.13  & 56.67  & 79.36  & 61.52  & 53.43  \\
    PFL~\cite{liu2022universal}   & 76.10  & 17.07  & 79.43  & 18.30  & 90.24  & 10.64  & 67.72  & 10.96  & 68.41  & \textbf{17.59} & 76.38  & 14.91  \\
    Hermes~\cite{gao2024training} & 57.82  & 61.09  & 74.27  & 35.45  & 87.56  & 16.91  & 50.97  & 39.42  & 49.24  & 76.55  & 63.97  & 45.89  \\
    AAL~\cite{chen2024versatile}   & 76.92  & 16.12  & 84.28  & 13.36  & 89.57  & 18.38  & 69.49  & 12.06  & 69.19  & 18.55  & 77.89  & 15.69  \\ \hline
    Ours  & \textbf{78.80} & \textbf{11.38} & \textbf{86.54} & \textbf{8.16} & \textbf{91.64} & \textbf{7.89} & \textbf{71.51} & \textbf{9.82} & \textbf{72.00} & 19.58  & \textbf{80.10} & \textbf{11.37} \\
    \bottomrule
    \end{tabular}
    }

\end{table}